\documentclass[11pt,a4paper]{article}
\usepackage[hyperref]{eacl2021}
\usepackage{times}
\usepackage{latexsym}

\usepackage{xcolor}
\usepackage{color, colortbl}
\usepackage{comment}
\usepackage{xspace}
\usepackage{multirow}
\usepackage{array}
\usepackage{float}
\usepackage{booktabs}
\usepackage{tikz}
\usepackage{caption}
\usepackage{enumitem}
\usepackage{tabu}
\usepackage{rotating}

\usepackage{microtype}

\aclfinalcopy %

\title{Putting Humans in the Natural Language Processing Loop: A Survey}

\author{
    Zijie J. Wang\Thanks{ denotes equal contribution}
    \quad Dongjin Choi\footnotemark[1]
    \quad Shenyu Xu\footnotemark[1]
    \quad Diyi Yang \vspace{2pt} \\
    College of Computing, Georgia Institute of Technology \vspace{2pt} \\
    \texttt{\{jayw, jin.choi, shenyuxu, dyang888\}@gatech.edu}
}

\date{}

\begin{document}
\maketitle

\definecolor{orange}{RGB}{255,119,0}
\definecolor{red}{RGB}{220,0,0}
\definecolor{agreen}{RGB}{74, 198, 148}
\definecolor{purple}{RGB}{158, 62, 177}
\definecolor{darkpurple}{RGB}{170, 70, 210}
\definecolor{aqua}{RGB}{87, 180, 181}
\definecolor{lightblue}{RGB}{72, 123, 232}
\definecolor{hotpink}{RGB}{255, 83, 115}
\definecolor{teal}{RGB}{90, 200, 250}
\definecolor{linkColor}{RGB}{6,125,233}
\definecolor{task}{HTML}{FC5C65}
\definecolor{goal}{HTML}{45AAF2}
\definecolor{interaction}{HTML}{FED330}
\definecolor{update}{HTML}{2BCBBA}

\definecolor{white}{HTML}{FFFFFF}
\definecolor{black}{HTML}{000000}
\definecolor{tabletagcolor}{HTML}{BDBDBD}

\definecolor{cell}{HTML}{B0BEC5}
\definecolor{rowbackground}{HTML}{F0F2F4}

\definecolor{cell}{HTML}{999999}
\definecolor{lightgray}{HTML}{F3F3F3}
\definecolor{customgray}{HTML}{585858}
\definecolor{rowbackground}{HTML}{F9F9F9}

\newcommand{\todo}[1]{\textcolor{hotpink}{[#1]}}
\newcommand{\jay}[1]{\textcolor{hotpink}{[#1 -jay]}}
\newcommand{\jin}[1]{\textcolor{aqua}{[#1 -jin]}}
\newcommand{\shenyu}[1]{\textcolor{orange}{[#1 -shenyu]}}
\newcommand{\diyi}[1]{\textcolor{agreen}{[#1 -diyi]}}

\newcommand{\mkclean}{
  \renewcommand{\jay}[1]{}
  \renewcommand{\jin}[1]{}
  \renewcommand{\shenyu}[1]{}
  \renewcommand{\diyi}[1]{}
  \renewcommand{\todo}[1]{}
}

\newcommand{\hitl}{HITL\xspace} %
\begin{abstract}

How can we design Natural Language Processing (NLP) systems that learn from human feedback?
There is a growing research body of Human-in-the-loop (\hitl) NLP frameworks that continuously integrate human feedback to improve the model itself.
\hitl NLP research is nascent but multifarious---solving various NLP problems, collecting diverse feedback from different people, and applying different methods to learn from collected feedback.
We present a survey of \hitl NLP work from both Machine Learning (ML) and Human-Computer Interaction (HCI) communities that highlights its short yet inspiring history, and thoroughly summarize recent frameworks focusing on their \textit{tasks}, \textit{goals}, \textit{human interactions}, and \textit{feedback learning methods}.
Finally, we discuss future directions for integrating human feedback in the NLP development loop.\looseness=-1

\end{abstract} %
\section{Introduction}

\begin{figure}[!tb]
    \includegraphics[width=\linewidth]{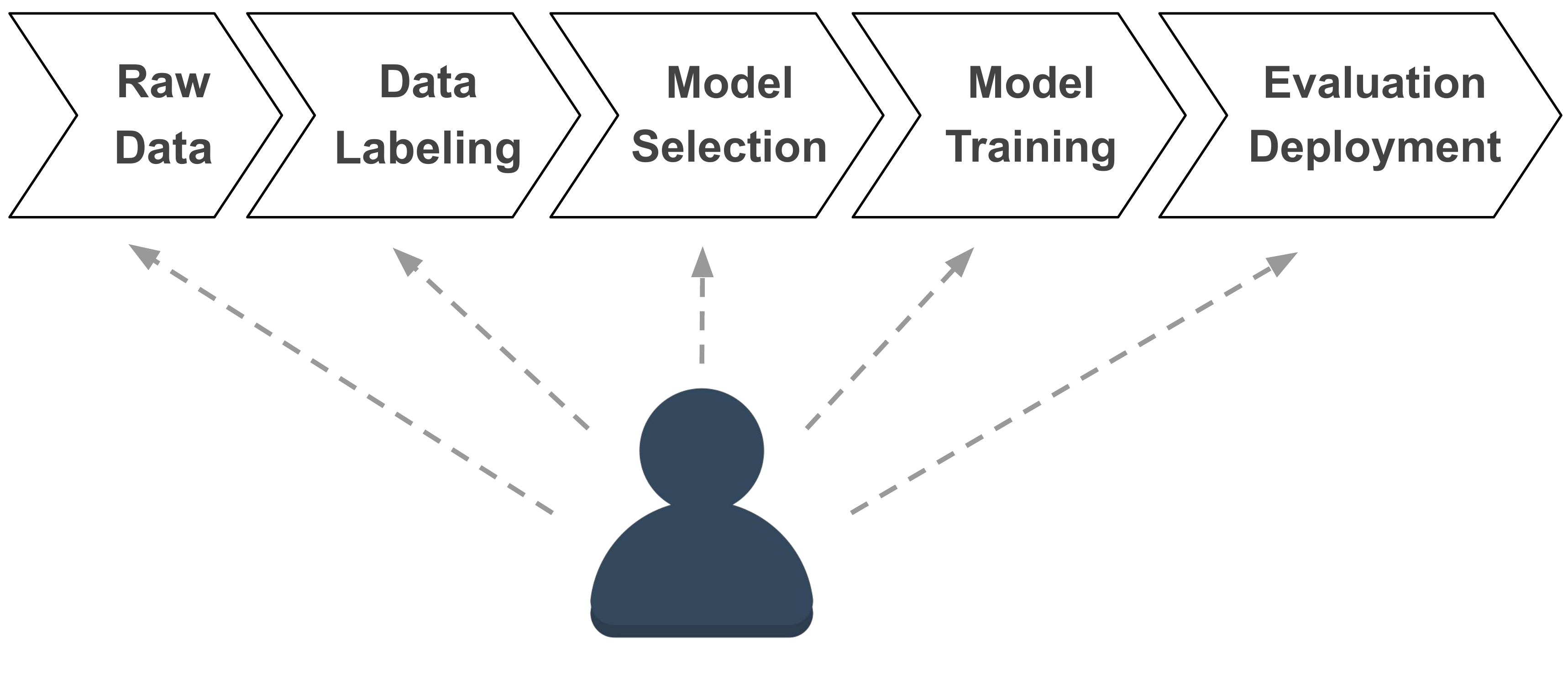}
    \caption{Collaboration between the human and the model in the Natural Language Processing Loop. The human provides feedback in different stages of the loop to improve the \textit{performance}, \textit{interpretability}, and \textit{usability} of NLP models.
    }
    \label{fig:HITLF}
\end{figure}

Traditionally, Natural Language Processing (NLP) models are trained, fine-tuned, and tested on existing dataset by machine learning experts, and then deployed to solve real-life problems of their users.
Model users can often give invaluable feedback that reveals design details overlooked by model developers, and provide data instances that are not represented in the training dataset \cite{kreutzerLearningHumanFeedback2020}.
However, the traditional linear NLP development pipeline is not designed to take advantage of human feedback.
Advancing on the conventional workflow, there is a growing research body of Human-in-the-loop (\hitl) NLP frameworks, or sometimes called mixed-initiative NLP, where model developers continuously integrate human feedback into different steps of the model deployment workflow (\autoref{fig:HITLF}).
This continuous feedback loop cultivates a human-AI partnership that not only enhances model accuracy and robustness, but also builds users' trust in NLP systems.

Just like traditional NLP frameworks, there is a high-dimensional design space for \hitl NLP systems. For example, human feedback can come from end users \cite{liDialogueLearningHumanInTheLoop2017} or crowd workers \cite{wallaceTrickMeIf2019}, and human can intervene models during training \cite{stiennonLearningSummarizeHuman2020} or deployment \cite{hancockLearningDialogueDeployment2019}.
Good \hitl NLP systems need to clearly \textit{communicates} to humans of what the model needs, provide intuitive \textit{interfaces} to collect feedback, and effectively \textit{learn} from them.
Therefore, \hitl NLP research spans across not only NLP and Machine Learning (ML) but also Human-computer Interaction (HCI).
A meta-analysis on existing \hitl NLP work focusing on bridging different research disciplines is vital to help new researchers quickly familiarize with this promising topic and recognize future research directions.
To fill this critical research gap, we provide a timely literature review on recent \hitl NLP studies from  both NLP and HCI communities.

This is the first comprehensive survey on the \hitl NLP topic. We make two main contributions: (1) We summarize recent studies on \hitl NLP and position each work with respect to its \textit{task}, \textit{goal}, \textit{human interaction}, and \textit{feedback learning method} (\autoref{table:papers}); (2) We critically discuss existing knowledge gaps and highlight important research directions that we distilled from the survey.
\section{Related Surveys}

To our best knowledge, there is no existing survey work on this topic, as the field for HITL NLP just starts growing.
Regarding interactive machine learning, related surveys like \citet{amershi2014power} focus on problems in interactive machine learning in general.
Specifically for reinforcement learning and dialogue system, there is one overview article~\cite{kreutzerLearningHumanFeedback2020} focusing on offline dialogue system.
Different from the aforementioned articles, our survey provides a comprehensive review on \hitl in NLP with critical discussions.

\newcolumntype{L}[1]{>{\raggedright\let\newline\\\arraybackslash\hspace{0pt}}m{#1}}
\newcolumntype{C}[1]{>{\centering\let\newline\\\arraybackslash\hspace{0pt}}m{#1}}
\newcolumntype{R}[1]{>{\raggedleft\let\newline\\\arraybackslash\hspace{0pt}}m{#1}}
\newcolumntype{P}[1]{>{\raggedright}p{#1}}

\newcommand{\f}{
\begin{tikzpicture}[every node/.style={inner sep=0,outer sep=0},scale=0.4]
    \fill [rounded corners=0.08cm,fill=cell] (0,0)--(.6,0)--(.6,.6)--(0,.6)--cycle;
\end{tikzpicture}
}

\newcommand{\rspace}{$\hspace{0.1cm}$}
\renewcommand{\arraystretch}{1.8}

\def \verticalAlignTwo {\hspace{1.0cm}\phantom{G}}
\def \tableRowSpace {0.15cm}

\begin{table*}[t]

\sffamily
\setlength{\tabcolsep}{0pt}
\renewcommand\arraystretch{1.30}

\centering
\scriptsize

\begin{tabular}{
R{3.4cm}|
C{0.55cm}C{0.55cm}C{0.55cm}C{0.55cm}C{0.55cm}|
C{0.55cm}C{0.55cm}C{0.55cm}|
C{0.55cm}C{0.55cm}C{0.55cm}C{0.55cm}C{0.55cm}C{0.55cm}C{0.55cm}|
C{0.55cm}C{0.55cm}C{0.55cm}
}

\multicolumn{1}{c}{} &
\multicolumn{5}{c}{{\small{\textbf{\textcolor{task}{TASK}}}}} &
\multicolumn{3}{c}{\small{\textbf{\textcolor{goal}{GOAL}}}} &
\multicolumn{7}{c}{\small{\textbf{\textcolor{interaction}{INTERACTION}}}} &
\multicolumn{3}{c}{\small{\textbf{\textcolor{update}{UPDATE}}}}
\\

\multicolumn{1}{r}{\textbf{Work} \rspace} & 
\multicolumn{1}{c}{\rotatebox{90}{Text Classification}} &
\multicolumn{1}{c}{\rotatebox{90}{Parsing and Entity Linking}} &
\multicolumn{1}{c}{\rotatebox{90}{Topic Modeling}} &
\multicolumn{1}{c}{\rotatebox{90}{Summarization and Machine Translation}} &
\multicolumn{1}{c|}{\rotatebox{90}{Dialogue and Question Answering System}} &
\multicolumn{1}{c}{\rotatebox{90}{Model Performance}} &
\multicolumn{1}{c}{\rotatebox{90}{Model Interpretability}} &
\multicolumn{1}{c|}{\rotatebox{90}{Usability}} &
\multicolumn{1}{c}{\rotatebox{90}{Mediums -- Graphical User Interface}} &
\multicolumn{1}{c}{\rotatebox{90}{Mediums -- Natural Language Interface}} &
\multicolumn{1}{c}{\rotatebox{90}{User Feedback Type -- Binary}} &
\multicolumn{1}{c}{\rotatebox{90}{User Feedback Type -- Scaled}} &
\multicolumn{1}{c}{\rotatebox{90}{User Feedback Type -- Natural Language}} &
\multicolumn{1}{c}{\rotatebox{90}{User Feedback Type -- Counterfactual Example}} &
\multicolumn{1}{c|}{\rotatebox{90}{Intelligent Interaction}} &
\multicolumn{1}{c}{\rotatebox{90}{Data Augmentation -- Offline Model Update}} &
\multicolumn{1}{c}{\rotatebox{90}{Data Augmentation -- Online Model Update}} &
\multicolumn{1}{c}{\rotatebox{90}{Model Direct Manipulation}}
\\
\midrule

\citet{godboleDocumentClassificationInteractive2004a} \rspace & \f & & & & & \f & & & \f & & \f & & & & \f & \f & & \\
\rowcolor{rowbackground}
\citet{settlesClosingLoopFast} \rspace & \f & & & & & \f & & & \f & & \f & & & & \f & \f & &\\
\citet{simardICEEnablingNonExperts}  \rspace & \f & & & & & \f & & & \f & & \f & \f & & & & \f & &\\
\rowcolor{rowbackground}
\citet{karmakharmJournalistintheLoopContinuousLearning2019} \rspace & \f & & & & & \f & & \f & \f & & \f & & & & & \f & & \\
\citet{jandotInteractiveSemanticFeaturing2016} \rspace & \f & & & & & & \f & & \f & & \f & & & & & \f & &\\
\rowcolor{rowbackground}
\citet{kaushik2019learning} \rspace & \f & & & & & \f & & & & \f & & & \f & \f & & \f & & \\
\citet{heHumanintheLoopParsing2016} \rspace & & \f & & & & \f & & & \f & & \f & & & & & \f & & \\
\rowcolor{rowbackground}
\citet{klieZeroHeroHumanInTheLoop2020} \rspace & & \f & & & & \f & & \f & \f & & \f & & & & & \f & & \\
\citet{loInteractiveEntityLinking2020} \rspace & & \f & & & & \f & & & \f & & \f & & & & \f & \f & & \\
\rowcolor{rowbackground}
\citet{trivedi2019interactive} \rspace & & \f & & & & \f & & & \f & & \f & & \f & & & \f & & \\
\citet{lawrence2018counterfactual} \rspace & & \f & & & & \f  & & & & \f & \f & & \f & \f & & \f & & \\
\rowcolor{rowbackground}
\citet{kimTopicSifterInteractiveSearch2019} \rspace & & & \f & & & \f & & \f & \f & & \f & & & & & & \f & \\
\citet{kumarWhyDidnYou2019a} \rspace & & & \f & & & \f & & & \f & & \f & \f & & & & & \f &\\
\rowcolor{rowbackground}
\citet{smithClosingLoopUserCentered2018} \rspace & & & \f & & & \f & \f & \f & \f & & \f & \f & & & & & \f &\\
\citet{stiennonLearningSummarizeHuman2020} \rspace & & & & \f& & \f & & & \f & & & \f & & & \f & \f & & \\
\rowcolor{rowbackground}
\citet{kreutzerCanNeuralMachine2018} \rspace & & & & \f & & \f & & & \f & & & \f & & & \f & & & \f \\
\citet{hancockLearningDialogueDeployment2019} \rspace & & & & & \f & \f & & & & \f & & & \f & & & & \f & \\
\rowcolor{rowbackground}
\citet{liuDialogueLearningHuman2018a} \rspace & & & & & \f & \f & & & & \f & \f & & \f & & \f & & \f & \f \\
\citet{liDialogueLearningHumanInTheLoop2017} \rspace & & & & & \f & \f & & & & \f & \f & & \f & & \f & \f & \f & \f \\
\rowcolor{rowbackground}
\citet{wallaceTrickMeIf2019} \rspace & & & & & \f & \f & & & & \f & \f & & \f & \f & & \f & & \\

\end{tabular}

\caption{
Overview of representative works in \hitl{} NLP.
Each row represents one work. Works are sorted by their task types. Each column corresponds to a dimension from the four subsections (task, goal, human interaction, and feedback learning method).
}

\label{table:papers}

\end{table*}
 
\section{Human-in-the-Loop NLP Tasks}
\label{sec:task}

This section describes what NLP sub-problems currently can benefit from a \hitl{} approach.
Over recent years, researchers and practitioners have made great advancements in NLP --- enabling many different NLP tasks and applications, such as text classification, summarization, and machine translation.
To effectively integrate humans into these task-specific development loops, people have developed many application-specific \hitl{} paradigms.
In this section, we categorize surveyed \hitl{} paradigms based on their corresponding tasks.

\subsection{Text Classification}

Text classification is a classic problem in NLP to categorize text into different groups.
Trained on a set of text documents ($X$) and their categorical labels ($Y$), a text classifier can predict $Y$ value of an unseen $X$.
For example, a movie review sentiment classifier can predict if one review is positive or negative.
Many \hitl{} frameworks are developed for this problem, where most of them start with \textit{training} a text classifier, then recruiting humans to \textit{annotate} data based on the current model behavior, and eventually \textit{retraining} the classifier on the larger dataset continuously.

For example, \citet{godboleDocumentClassificationInteractive2004a} develop a \hitl{} paradigm where users can interactively edit (add or remove) text features and label new documents.
Also, \citeauthor{godboleDocumentClassificationInteractive2004a} integrate active learning in their framework---instead of arbitrarily presenting data for users to annotate, they strategically choose samples that can maximize the expected information gain.
With active learning, labelers can annotate fewer data to achieve the same model improvement of a framework using random sampling.
\citet{settlesClosingLoopFast} further improves this \hitl{} workflow by extending the active learning component to features (words) sampling in addition to labels (documents).
To ease the deployment of \hitl{} text classifiers, \citet{simardICEEnablingNonExperts} develop a robust web-based tool that supports general text classification tasks.
Researchers have also developed domain-specific \hitl{} text classification systems.
For instance, a rumor classification system developed for journalists tightly integrates model retraining, data collection, and data annotation in deployment \cite{karmakharmJournalistintheLoopContinuousLearning2019}.

Bag-of-words feature is commonly used in text classifiers.
However, \citet{jandotInteractiveSemanticFeaturing2016} explore alternative feature representations to better support a \hitl{} pipeline.
A dictionary, also called lexicon and gazetteer, is a set of words sharing the same semantics.
Well-defined dictionaries give higher accuracy and are more interpretable than bag-of-words.
In \citeauthor{jandotInteractiveSemanticFeaturing2016}’s \hitl{} system, users can easily create semantic dictionaries through a web-based user interface, and the text classifier is continuously trained on new dictionaries.

\subsection{Parsing and Entity Linking}

Besides classifying documents, recent research shows great potential of \hitl{} approach in enhancing the performance of existing parsing and entity linking models.
Parsing in NLP is a process to determine the syntactic structure of the input text.
Entity linking aims to assign unique identity to entities in the text, such as names and locations.

Advancing traditional Combinatory Categorial Grammars (CCG) parsers, \citet{heHumanintheLoopParsing2016} crowdsource parsing tasks---a trained parser is uncertain about---to non-expert mechanical turks, by asking them simple what-questions.
Using human feedback as a soft constraint to penalize the parser during retraining, the performance of the original parser improves significantly.
Similarly, \citet{klieZeroHeroHumanInTheLoop2020} recruit humans to interactively annotate correct entities in text samples where an entity linking model performs poorly.
Also, with more strategic sampling methods to select instances to present to humans, a smaller set of feedback can quickly improve the entity linking model performance \cite{klieZeroHeroHumanInTheLoop2020, loInteractiveEntityLinking2020}.

\subsection{Topic Modeling}

In addition to using a \hitl{} approach to enhance learning the low-level semantic relationships, researchers apply similar frameworks to topic modeling techniques that are used to analyze large document collections.
People traditionally use \textit{take it or leave it} algorithms for this task, such as information retrieval and document clustering.
Over the past few years, there is a growing research body of \hitl{} topic modeling \cite{leeHumanTouchHow2017}.
For example, \citet{huInteractiveTopicModeling2014}'s and \citet{jaegulchooUTOPIANUserDrivenTopic2013}'s systems allow users to refine a trained model through adding, removing, or changing the weights of words within each topic.
Then, using the user-updated features and weights, trained models are more likely to generate useful topics.

More recently, researchers focus on developing more human-centered \hitl{} topic modeling methods.
These methods emphasize the needs of topic modeling end users, mostly NLP non-experts, instead of only collecting algorithmically convenient human feedback \cite{leeHumanTouchHow2017}.
For example, \citet{kimTopicSifterInteractiveSearch2019} develop an intuitive visualization system that allows end users to up-vote or down-vote specific documents to inform their interest to the model.
\citet{smithClosingLoopUserCentered2018} conduct users studies with non-experts and develop a responsive and predictable user interface that supports a broad range of topic modeling refinement operations.
These examples show that \hitl{} NLP systems can benefit from HCI design techniques.

\vspace{-1mm}
\subsection{Summarization and Machine Translation}

Besides using a \hitl{} approach to analyze existing documents, researchers also apply them to generate new texts.
Text summarization and machine translation have seen major breakthroughs in recent years \cite{brownLanguageModelsAre2020}, which draw attentions from both NLP and HCI communities to design and develop \hitl{} systems.
For example, \citet{stiennonLearningSummarizeHuman2020} collect human preferences on pairs of summaries generated by two models, then train a reward model to predict the preference.
Then, this reward model is used to train a policy to generate summaries using reinforcement learning.
Similarly, \citet{kreutzerCanNeuralMachine2018} collect both explicit and implicit human feedback to improve a machine translation model by using the feedback with reinforcement learning.
Experiments show that the model trained on human preference data has higher accuracy and better generalization.

\subsection{Dialogue and Question Answering}
\label{subsec:qa-system}

Recently, many HITL frameworks have been developed for dialogue and Question Answering (QA) systems, where the AI agent can have conversation with users.
We can group these systems into two categories: (1) \textit{online feedback loop}: the system continuously uses human feedback to update the model; (2) \textit{offline feedback loop}: human feedback is filtered and aggregated to update the model in batch \cite{kreutzerLearningHumanFeedback2020}.

\textbf{Online feedback loop.} 
Traditionally, there is a mismatch of the training set and online use case for dialogue systems.
To tackle this challenge, online reinforcement learning can be used to improve the model with human feedback.
For example, \citet{liuDialogueLearningHuman2018a} collect dialogue corrections from users during deployment, while \citet{liDialogueLearningHumanInTheLoop2017} collect both binary explicit feedback and implicit natural language feedback.
Also, \citet{hancockLearningDialogueDeployment2019} propose a lifetime learning framework to improve chatbot performance.
The chatbot is trained not only to generate dialogues but also to predict user satisfactions.
During deployment, the chatbot predicts user satisfaction after generating responses, and asks for user feedback if the predicted satisfaction score is low.
Then, the chatbot uses the feedback as a new training example to retrain itself continuously.\looseness=-1

\textbf{Offline feedback loop.}
With offline \hitl{}, model is updated after collecting a large set of human feedback.
For instance, \citet{wallaceTrickMeIf2019} invite crowd workers to generate adversarial questions that can fool their QA system, and use these questions for adversarial training.
Offline feedback loop can be more robust for dialogue systems, because user feedback can be misleading, so directly updating the model is risky \cite{kreutzerLearningHumanFeedback2020}.

\section{Human-in-the-Loop Goals}
\label{sec:goal}

Among surveyed papers, the most common motivation for using a \hitl{} approach in NLP tasks is to improve the \textbf{model performance}.
There are different metrics to measure model performance, and experiments in our surveyed papers show that \hitl{} can significantly and effectively improve model performance with a relatively small set of human feedback.
For example, for text classification, \hitl{} improves classification accuracy \cite{smithClosingLoopUserCentered2018, jandotInteractiveSemanticFeaturing2016}.
Similarly, dialogue and question answering systems have higher ranking metric hits after adapting a \hitl{} approach \cite{hancockLearningDialogueDeployment2019, brownLanguageModelsAre2020}.
Researchers also find \hitl{} improves model's robustness and generalization on different data \cite{stiennonLearningSummarizeHuman2020, jandotInteractiveSemanticFeaturing2016}.\looseness=-1

In addition to model performance, \hitl{} can also improve the \textbf{interpretability} and \textbf{usability} of NLP models.
For example, \citet{jandotInteractiveSemanticFeaturing2016} enable users to create semantic dictionaries.
These features have semantic meanings and are more interpretable to humans.
\citet{wallaceTrickMeIf2019} guide humans to generate model-specific adversarial questions that can fool the question answering model.
These adversarial questions can be used as probes to gain insights of the underlying model behaviors.
On the other hand, \hitl{} can also improve the user experience with NLP models.
For example, \citet{smithClosingLoopUserCentered2018} develop an interface for topic modeling users to intuitively interact and control their models.
User studies show that users gain more trust and confidence through the \hitl{} system.

\section{Human-machine Interaction}
\label{sec:humanmachineinterface}

This section discusses the mediums that users use to interact with HITL systems and different types of feedback that the systems collect.
This section is strongly correlated with \autoref{sec:usefeedback}, which describes how existing works leverage user feedback to update models.
By first describing how and what user feedback may be collected (this section), we can more easily ground our discussion on how to leverage the feedback (next section).

\subsection{Interaction Mediums}
There are two common interaction mediums shared by our surveyed systems: \textit{graphical user interface} and \textit{natural language interface}.

\subsubsection{Graphical User Interface}
One of the commonly used interaction mediums for collecting user feedback is the Graphical User Interface (GUI).
The GUI provides a user interface that allows users to interact with systems through graphical icons and visual indicators such as secondary notations.
Some \hitl NLP systems allow users to directly label samples in the GUI~\cite{simardICEEnablingNonExperts,godboleDocumentClassificationInteractive2004a,settlesClosingLoopFast}.
The GUI also makes feature editing possible for end-users who do not develop the model from initial~\cite{jandotInteractiveSemanticFeaturing2016,simardICEEnablingNonExperts,godboleDocumentClassificationInteractive2004a}.
Some other works even use the GUI for users to rate training sentences in the text summarization task~\cite{stiennonLearningSummarizeHuman2020} and rank the generated topics in the topic modeling task~\cite{kimTopicSifterInteractiveSearch2019}.
One obvious advantage of the GUI is that it visualizes the NLP model running in the black box, enhancing the interpretability of the model as seen in \autoref{sec:goal}.
In addition, compared to text-based user interfaces, the GUI supports Windows, Icons, Menus, Pointer (WIMP) interactions, providing users more accurate control for refining the model.

\subsubsection{Natural Language Interface}
Another commonly used interaction medium in \hitl NLP systems is natural language interface.
A natural language interface is an interface where the user interacts with the computer through natural language.
As this interface usually simulates having a conversation with a computer, it mostly comes with the purpose of building up a dialogue system~\cite{hancockLearningDialogueDeployment2019,liuDialogueLearningHuman2018a,liDialogueLearningHumanInTheLoop2017}.
The natural language interface not only supports users to provide explicit feedback~\cite{liuDialogueLearningHuman2018a,liDialogueLearningHumanInTheLoop2017}, such as positive or negative responses.
It also allows users to give implicit feedback with natural language sentences~\cite{hancockLearningDialogueDeployment2019,liDialogueLearningHumanInTheLoop2017}.
Compared to the GUI, the natural language interface is more intuitive to use as it simulates the process of human’s conversation and thus needs no additional tutorial.
It can also be perfectly integrated with the dialogue system and supports the collection of natural language feedback, providing users more freedom for refining the model, as discussed in \autoref{sec:usefeedback}.

\subsection{User Feedback Types}
\label{subsec:feedback-types}
Above, we discussed that the GUI and the natural language interface are two common interaction mediums implemented in the \hitl systems we surveyed for collecting user feedback.
In the following, we will discuss four major types of user feedback supported by the two interaction mediums, including binary user feedback, scaled user feedback, natural language user feedback, and counterfactual example feedback.

\subsubsection{Binary User Feedback}
\label{subsubsec:binary-feedback}
Binary user feedback is the feedback which has two categories that are usually opposite to each other, such as ``like'' and ``dislike''.
It can be collected by both the GUI and the natural language interface.
As discussed above, the GUI can collect binary user feedback from the user's adding or removing labels~\cite{simardICEEnablingNonExperts}
and features~\cite{godboleDocumentClassificationInteractive2004a}.
The natural language interface can also support binary user feedback collection with simple short natural language response, such as ``agree’’ or ``reject’’~\cite{liuDialogueLearningHuman2018a,liDialogueLearningHumanInTheLoop2017}.
As binary user feedback only contains two categories, this kind of user feedback is usually easy to collect but sometimes may over-simplify users' intention by ignoring the potential intermediate situations.
Binary user feedback can be used to provide explicit feedback for the system to update training datasets or directly manipulate models, as discussed in \autoref{sec:usefeedback}.

\begin{table*}
\sffamily
\centering
\small
\begin{tabular}{p{1.8cm}p{4.6cm}p{4cm}p{3.6cm}}
&
Offline Model Update 
&
Online Model Update 
&
Model Direct Manipulation
\\ \hline
Binary & \citet{klieZeroHeroHumanInTheLoop2020},~\citet{loInteractiveEntityLinking2020},~\citet{trivedi2019interactive},~\citet{karmakharmJournalistintheLoopContinuousLearning2019},~\citet{wallaceTrickMeIf2019}, \citet{godboleDocumentClassificationInteractive2004a}, etc.
&
\citet{kimTopicSifterInteractiveSearch2019},~\citet{kumarWhyDidnYou2019a},~\citet{smithClosingLoopUserCentered2018},~\citet{liuDialogueLearningHuman2018a},~\citet{liDialogueLearningHumanInTheLoop2017}
& 
\citet{liuDialogueLearningHuman2018a},~\citet{liDialogueLearningHumanInTheLoop2017}\\ 
Scaled & \citet{stiennonLearningSummarizeHuman2020},~\citet{simardICEEnablingNonExperts}     & 
\citet{kumarWhyDidnYou2019a},~\citet{smithClosingLoopUserCentered2018}
&  \citet{kreutzerCanNeuralMachine2018}    \\
Natural \newline Language
& \citet{kaushik2019learning},~\citet{trivedi2019interactive},~\citet{wallaceTrickMeIf2019},~\citet{lawrence2018counterfactual},~\citet{liDialogueLearningHumanInTheLoop2017}     & 
\citet{hancockLearningDialogueDeployment2019},~\citet{liuDialogueLearningHuman2018a},~\citet{liDialogueLearningHumanInTheLoop2017}
 &  \citet{liuDialogueLearningHuman2018a},~\citet{liDialogueLearningHumanInTheLoop2017}    \\
Counterfactual Example & \citet{kaushik2019learning},~\citet{wallaceTrickMeIf2019},~\citet{lawrence2018counterfactual} & --- & ---    \\
\end{tabular}

\caption{
Relationship between the user feedback types and how they are used.
Each row represents one feedback type (\autoref{subsec:feedback-types}), and each column corresponds to a model learning method (\autoref{sec:usefeedback}).
}

\label{table:papers}

\end{table*}
 
\subsubsection{Scaled User Feedback}
Scaled user feedback is the feedback which has scaled categories and is usually in numerical formats, such as the 5-point scale rating.
It often can only be collected through the GUI as it is difficult to express accurate scaled feedback in natural language.
Such user feedback is collected in the GUI when users rate their preferences of training data or model results~\cite{kreutzerCanNeuralMachine2018}
and adjust features on a numerical scale~\cite{simardICEEnablingNonExperts}.
Similar to binary user feedback, scaled user feedback can provide explicit feedback for the system to update the models but with more options to cover intermediate cases (e.g. adjusting the weight of one feature from 1 to 3 on a scale of 5 points).
Besides, the scaled ratings of user preferences can also be used as implicit guidance for the system to improve the model, as discussed in \autoref{subsec:dataupdate} and \autoref{subsec:model-manipulation}.

\subsubsection{Natural Language User Feedback}
Compared with binary user feedback and scaled user feedback, natural language user feedback is a better way for representing users' intention but vague and hard for the machine to interpret because of the intuitive property of human language.
As mentioned above, natural language user feedback can only be collected through the natural language interface.
Users provide this type of user feedback by directly inputting natural language text into the system~\cite{hancockLearningDialogueDeployment2019,liDialogueLearningHumanInTheLoop2017}.
By analyzing the user input sentences, the system implies the user’s intention and accordingly updates the model, as seen in \autoref{subsec:dataupdate} and \autoref{subsec:model-manipulation}.\looseness=-1

\subsection{Counterfactual Example Feedback}
Similar to the natural language user feedback, counterfactual example feedback is usually in the form of natural language text and collected through the natural language interface.
A counterfactual example describes a causal situation in the form: "If X had not occurred, Y would not have occurred."
The \hitl NLP systems collect and analyze user-modified counterfactual text examples and retrain the model accordingly~\cite{kaushik2019learning, lawrence2018counterfactual}, as more details covered in \autoref{subsec:dataupdate}.

\subsection{Intelligent Interaction}
\label{subsec:intelligent}
In addition to the choice of the interaction medium and the collected user feedback types, some \hitl NLP systems also leverage intelligent interaction techniques to enhance human-machine interaction.
As discussed in \autoref{sec:task}, \textbf{active learning} and \textbf{reinforcement learning} are two commonly used techniques we observed in our surveyed systems.
Active learning allows the system to interactively query a user to label new data points with the desired outputs~\cite{godboleDocumentClassificationInteractive2004a, settlesClosingLoopFast, loInteractiveEntityLinking2020}.
By strategically choosing samples to maximize information gain with fewer iterations, active learning not only reduces human efforts on data labeling but also improves the efficiency of the system.
Compared to active learning, reinforcement learning takes actions based on user feedback to maximize the notion of cumulative reward~\cite{stiennonLearningSummarizeHuman2020, kreutzerCanNeuralMachine2018, liDialogueLearningHumanInTheLoop2017, liuDialogueLearningHuman2018a}.
By considering each user feedback as a new action, reinforcement learning supports accurate understanding of human intention and updating the model accordingly, as discussed in \autoref{sec:usefeedback}.
\section{How to Use User Feedback}
\label{sec:usefeedback}

This section summarizes how existing \hitl NLP systems utilize feedback.
Different feedback types described in~\autoref{subsec:feedback-types} are leveraged by the systems with different update methods (\autoref{table:papers}).
In the following, we will discuss two major update methods, including \textit{data augmentation} and \textit{model direct manipulation}.

\subsection{Data Augmentation}
\label{subsec:dataupdate}
One popular approach is to consider the feedback as a new ground truth data sample.
For example, a user’s answer to a model’s question can be a data sample to retrain a QA model.
We describe two types of techniques to use augmented data set:
\textit{Offline update} retrains NLP model from scratch after collecting human feedback, while \textit{Online update} trains NLP models while collecting feedback at the same time.

\textbf{Offline model update}
is usually performed after certain amount of human feedback is collected.
Offline update does not need to be immediate, so it is suitable for noisy feedback with complex models which takes extra processing and training time.
For example, \citet{simardICEEnablingNonExperts} and \citet{karmakharmJournalistintheLoopContinuousLearning2019} use human feedback as new class labels and span-level annotations, and retrain their models after collecting enough new data. 

\textbf{Online model update}
is applied right after user feedback is given.
This is effective for dialogue systems and conversational QA systems where recent input is crucial to machine's reasoning \cite{liDialogueLearningHumanInTheLoop2017}.
\textit{Incremental learning} technique is often used to learn augmented data in real-time \cite{kumarWhyDidnYou2019a}.
It focuses on making an incremental change to current system using the newly come feedback information effectively.
Interactive topic modeling systems and feature engineering systems widely use this technique.
For example, \citet{kimTopicSifterInteractiveSearch2019} incrementally updates topic hierarchy by extending or shrinking topic tree incrementally.
Also, some frameworks use the latent Dirichlet allocation (LDA) to adjust sampling parameters with collected feedback in incremental iterations~\cite{smithClosingLoopUserCentered2018}.\looseness=-1

\subsection{Model Direct Manipulation}
\label{subsec:model-manipulation}

Collected numerical human feedback are usually directly used to adjust model's objective function.
For example, \citet{liDialogueLearningHumanInTheLoop2017}
collect binary feedback as rewards for reinforcement learning of a dialogue agent.
Similarly, \citet{kreutzerCanNeuralMachine2018} use a 5-point scale rating as reward function of reinforcement and bandit learning for machine translation.
Existing works have focused more on numerical feedback than natural language feedback.
Numerical feedback is easier to be incorporated into models, but provides limited information compared to natural language.
For future research, incorporating more types of feedback (e.g., speech, log data) will be an interesting direction to gain more useful insights from humans.
With more complex feedback types, it is critical to design both quantitative and qualitative methods to evaluate collected feedback, as they can be noisy just like any other data.
\section{Research Directions \& Open Problems}

In this section, we highlight research directions and open problems, which are distilled from the surveyed papers, for future research.

\subsection{Broader Roles of \hitl NLP System}

Improving model performance is the most popular goal among surveyed NLP \hitl{} frameworks.
However, researchers have found \hitl{} method also enhances NLP model interpretability \cite{wallaceTrickMeIf2019} and usability \cite{leeHumanTouchHow2017}.
Therefore, we encourage future NLP researchers to explore \hitl{} as a mean to better understand their models and improve the user experience of model end users.
For example, user feedback can be used to mitigate harms caused by NLP model bias~\cite{blodgett-etal-2020-language}.
While many recent models gain feedback from \textit{mechanic turks} \cite{heHumanintheLoopParsing2016}, future researchers can consider involving \textit{model engineers} and \textit{end users} in their NLP development loop \cite{hancockLearningDialogueDeployment2019}. 
For example, one can develop a web-based tool where end users can interactively modify the feature weights of a text classifier and observe the model behavior at run time.
By performing these ``what-if'' operations, users can gain additional insights of how model internally uses these features.
Similarly, a \hitl{} chatbot could grant users more control by supporting model parameter modification through natural language input.
For instance, a non-binary user Alex could correct the chatbot's pronoun use by typing ``Hello chatbot, could you use they/them/theirs to refer me in the future please?"

\subsection{Human-centered System Design}

In this survey, we found most of the \hitl NLP systems and techniques are designed and developed by NLP researchers.
We believe that this area of research will be greatly benefited from a deeper involvement of the HCI community.
Human feedback is the key for \hitl{} systems.
However, with a poorly designed human-machine interface, the collected human feedback are more likely to be inconsistent, incorrect, or even misleading.
Therefore, better interface design and user studies to evaluate \hitl{} system interface can greatly enhance the quality of feedback collection, which in turn improves the downstream task performance.

To shed light on \hitl{} NLP research from a HCI perspective, \citet{wallaceTrickMeIf2019} explore the effect of adding model interpretation cues in the \hitl{} interface on the quality of collected feedback;
\citet{schochThisProblem2020} investigate the impacts of question framing imposed on humans;
similarly, \citet{raoLearningAskGood2018} study how to ask good questions to which humans are more likely to give helpful feedback.
There are many exciting research opportunities and challenges in designing and evaluating \hitl{} interface.
As a starting point for developing more human-centered \hitl{} NLP systems, we provide the following concrete research directions:

\begin{itemize}[topsep=2mm, itemsep=0.6mm, parsep=0mm, leftmargin=4mm, label=$\triangleright$]
    \item As human feedback can be subjective, who should \hitl{} NLP systems collect feedback from? Is there any expertise levels required to perform certain tasks~\cite{kreutzerLearningHumanFeedback2020}?
    \item How to present what the model has learned and what feedback is needed? How to visualize the model change after learning from user feedback~\cite{leeHumanTouchHow2017}?
    \item How to dynamically choose the most helpful feedback to collect~\cite{settlesClosingLoopFast}? How to guide users to provide useful feedback~\cite{wallaceTrickMeIf2019}?
    \item How to evaluate the collected human feedback as it can be noisy and even misleading~\cite{kreutzerLearningHumanFeedback2020}?
    \item Conduct rigorous user studies to evaluate the effectiveness of \hitl{} systems in addition to model performance~\cite{smithClosingLoopUserCentered2018}.
    \item Open-source tools and share user study protocols when publishing new \hitl{} NLP work.
    \item Create and share human feedback datasets.
\end{itemize}

\section{Conclusion}
In this paper, we conduct a comprehensive survey on \hitl NLP.
We summarize recent literature on \hitl NLP from both NLP and HCI communities, and position each work with respect to its \textit{task}, \textit{goal}, \textit{human interaction}, and \textit{feedback learning method}.
The field of \hitl NLP is still relatively nascent, and we see very diverse system design methods.
To help new researchers and practitioners quickly familiarize with the field, we recognize the great potential of \hitl NLP systems in different NLP tasks and highlight open challenges and future research directions.
The research directions rest on a greater collaboration between NLP and HCI researchers and practitioners---a paramount step to create next-generation NLP technologies that deeply align with people's needs. 
\bibliography{survey}

\begin{thebibliography}{29}
\expandafter\ifx\csname natexlab\endcsname\relax\def\natexlab#1{#1}\fi

\bibitem[{Amershi et~al.(2014)Amershi, Cakmak, Knox, and
  Kulesza}]{amershi2014power}
Saleema Amershi, Maya Cakmak, William~Bradley Knox, and Todd Kulesza. 2014.
\newblock Power to the people: The role of humans in interactive machine
  learning.
\newblock \emph{Ai Magazine}, 35(4):105--120.

\bibitem[{Blodgett et~al.(2020)Blodgett, Barocas, Daum{\'e}~III, and
  Wallach}]{blodgett-etal-2020-language}
Su~Lin Blodgett, Solon Barocas, Hal Daum{\'e}~III, and Hanna Wallach. 2020.
\newblock \href {https://doi.org/10.18653/v1/2020.acl-main.485} {Language
  (technology) is power: A critical survey of {``}bias{''} in {NLP}}.
\newblock In \emph{Proceedings of the 58th Annual Meeting of the Association
  for Computational Linguistics}, pages 5454--5476, Online. Association for
  Computational Linguistics.

\bibitem[{Brown et~al.(2020)Brown, Mann, Ryder, Subbiah, Kaplan, Dhariwal,
  Neelakantan, Shyam, Sastry, Askell, Agarwal, {Herbert-Voss}, Krueger,
  Henighan, Child, Ramesh, Ziegler, Wu, Winter, Hesse, Chen, Sigler, Litwin,
  Gray, Chess, Clark, Berner, McCandlish, Radford, Sutskever, and
  Amodei}]{brownLanguageModelsAre2020}
Tom~B. Brown, Benjamin Mann, Nick Ryder, Melanie Subbiah, Jared Kaplan,
  Prafulla Dhariwal, Arvind Neelakantan, Pranav Shyam, Girish Sastry, Amanda
  Askell, Sandhini Agarwal, Ariel {Herbert-Voss}, Gretchen Krueger, Tom
  Henighan, Rewon Child, Aditya Ramesh, Daniel~M. Ziegler, Jeffrey Wu, Clemens
  Winter, Christopher Hesse, Mark Chen, Eric Sigler, Mateusz Litwin, Scott
  Gray, Benjamin Chess, Jack Clark, Christopher Berner, Sam McCandlish, Alec
  Radford, Ilya Sutskever, and Dario Amodei. 2020.
\newblock \href {http://arxiv.org/abs/2005.14165} {Language {{Models}} are
  {{Few}}-{{Shot Learners}}}.
\newblock \emph{arXiv:2005.14165 [cs]}.

\bibitem[{Godbole et~al.(2004)Godbole, Harpale, Sarawagi, and
  Chakrabarti}]{godboleDocumentClassificationInteractive2004a}
Shantanu Godbole, Abhay Harpale, Sunita Sarawagi, and Soumen Chakrabarti. 2004.
\newblock Document {{Classification Through Interactive Supervision}} of
  {{Document}} and {{Term Labels}}.
\newblock In \emph{Knowledge {{Discovery}} in {{Databases}}: {{PKDD}} 2004},
  volume 3202, pages 185--196. {Springer Berlin Heidelberg}, {Berlin,
  Heidelberg}.

\bibitem[{Hancock et~al.(2019)Hancock, Bordes, Mazar{\'e}, and
  Weston}]{hancockLearningDialogueDeployment2019}
Braden Hancock, Antoine Bordes, Pierre-Emmanuel Mazar{\'e}, and Jason Weston.
  2019.
\newblock \href {http://arxiv.org/abs/1901.05415} {Learning from {{Dialogue}}
  after {{Deployment}}: {{Feed Yourself}}, {{Chatbot}}!}
\newblock \emph{arXiv:1901.05415 [cs, stat]}.

\bibitem[{He et~al.(2016)He, Michael, Lewis, and
  Zettlemoyer}]{heHumanintheLoopParsing2016}
Luheng He, Julian Michael, Mike Lewis, and Luke Zettlemoyer. 2016.
\newblock Human-in-the-{{Loop Parsing}}.
\newblock In \emph{Proceedings of the 2016 {{Conference}} on {{Empirical
  Methods}} in {{Natural}} {{Language Processing}}}, pages 2337--2342, {Austin,
  Texas}. {Association for Computational Linguistics}.

\bibitem[{Hu et~al.(2014)Hu, {Boyd-Graber}, Satinoff, and
  Smith}]{huInteractiveTopicModeling2014}
Yuening Hu, Jordan {Boyd-Graber}, Brianna Satinoff, and Alison Smith. 2014.
\newblock Interactive topic modeling.
\newblock \emph{Machine Learning}, 95(3):423--469.

\bibitem[{{Jaegul Choo} et~al.(2013){Jaegul Choo}, {Changhyun Lee}, Reddy, and
  Park}]{jaegulchooUTOPIANUserDrivenTopic2013}
{Jaegul Choo}, {Changhyun Lee}, Chandan~K. Reddy, and Haesun Park. 2013.
\newblock {{UTOPIAN}}: {{User}}-{{Driven Topic Modeling Based}} on
  {{Interactive Nonnegative Matrix Factorization}}.
\newblock \emph{IEEE Transactions on Visualization and Computer Graphics},
  19(12):1992--2001.

\bibitem[{Jandot et~al.(2016)Jandot, Simard, Chickering, Grangier, and
  Suh}]{jandotInteractiveSemanticFeaturing2016}
Camille Jandot, Patrice Simard, Max Chickering, David Grangier, and Jina Suh.
  2016.
\newblock \href {http://arxiv.org/abs/1606.07545} {Interactive {{Semantic
  Featuring}} for {{Text Classification}}}.
\newblock \emph{arXiv:1606.07545 [cs, stat]}.

\bibitem[{Karmakharm et~al.(2019)Karmakharm, Aletras, and
  Bontcheva}]{karmakharmJournalistintheLoopContinuousLearning2019}
Twin Karmakharm, Nikolaos Aletras, and Kalina Bontcheva. 2019.
\newblock Journalist-in-the-{{Loop}}: {{Continuous Learning}} as a {{Service}}
  for {{Rumour Analysis}}.
\newblock In \emph{Proceedings of the 2019 {{Conference}} on {{Empirical
  Methods}} in {{Natural Language Processing}} and the 9th {{International
  Joint Conference}} on {{Natural Language Processing}} ({{EMNLP}}-{{IJCNLP}}):
  {{System Demonstrations}}}, pages 115--120, {Hong Kong, China}. {Association
  for Computational Linguistics}.

\bibitem[{Kaushik et~al.(2019)Kaushik, Hovy, and Lipton}]{kaushik2019learning}
Divyansh Kaushik, Eduard Hovy, and Zachary~C Lipton. 2019.
\newblock Learning the difference that makes a difference with
  counterfactually-augmented data.
\newblock \emph{arXiv preprint arXiv:1909.12434}.

\bibitem[{Kim et~al.(2019)Kim, Choi, Drake, Endert, and
  Park}]{kimTopicSifterInteractiveSearch2019}
Hannah Kim, Dongjin Choi, Barry Drake, Alex Endert, and Haesun Park. 2019.
\newblock {{TopicSifter}}: {{Interactive Search Space Reduction}} through
  {{Targeted Topic Modeling}}.
\newblock In \emph{2019 {{IEEE Conference}} on {{Visual Analytics Science}} and
  {{Technology}} ({{VAST}})}, pages 35--45, {Vancouver, BC, Canada}. {IEEE}.

\bibitem[{Klie et~al.(2020)Klie, {Eckart de Castilho}, and
  Gurevych}]{klieZeroHeroHumanInTheLoop2020}
Jan-Christoph Klie, Richard {Eckart de Castilho}, and Iryna Gurevych. 2020.
\newblock From {{Zero}} to {{Hero}}: {{Human}}-{{In}}-{{The}}-{{Loop Entity
  Linking}} in {{Low Resource Domains}}.
\newblock In \emph{Proceedings of the 58th {{Annual Meeting}} of the
  {{Association}} for {{Computational Linguistics}}}, pages 6982--6993,
  {Online}. {Association for Computational Linguistics}.

\bibitem[{Kreutzer et~al.(2018)Kreutzer, Khadivi, Matusov, and
  Riezler}]{kreutzerCanNeuralMachine2018}
Julia Kreutzer, Shahram Khadivi, Evgeny Matusov, and Stefan Riezler. 2018.
\newblock \href {http://arxiv.org/abs/1804.05958} {Can {{Neural Machine
  Translation}} be {{Improved}} with {{User Feedback}}?}
\newblock \emph{arXiv:1804.05958 [cs, stat]}.

\bibitem[{Kreutzer et~al.(2020)Kreutzer, Riezler, and
  Lawrence}]{kreutzerLearningHumanFeedback2020}
Julia Kreutzer, Stefan Riezler, and Carolin Lawrence. 2020.
\newblock \href {http://arxiv.org/abs/2011.02511} {Learning from {{Human
  Feedback}}: {{Challenges}} for {{Real}}-{{World Reinforcement Learning}} in
  {{NLP}}}.
\newblock \emph{arXiv:2011.02511 [cs]}.

\bibitem[{Kumar et~al.(2019)Kumar, {Smith-Renner}, Findlater, Seppi, and
  {Boyd-Graber}}]{kumarWhyDidnYou2019a}
Varun Kumar, Alison {Smith-Renner}, Leah Findlater, Kevin Seppi, and Jordan
  {Boyd-Graber}. 2019.
\newblock Why {{Didn}}'t {{You Listen}} to {{Me}}? {{Comparing User Control}}
  of {{Human}}-in-the-{{Loop Topic Models}}.
\newblock In \emph{Proceedings of the 57th {{Annual Meeting}} of the
  {{Association}} for {{Computational Linguistics}}}, pages 6323--6330,
  {Florence, Italy}. {Association for Computational Linguistics}.

\bibitem[{Lawrence and Riezler(2018)}]{lawrence2018counterfactual}
Carolin Lawrence and Stefan Riezler. 2018.
\newblock Counterfactual learning from human proofreading feedback for semantic
  parsing.
\newblock \emph{arXiv preprint arXiv:1811.12239}.

\bibitem[{Lee et~al.(2017)Lee, Smith, Seppi, Elmqvist, {Boyd-Graber}, and
  Findlater}]{leeHumanTouchHow2017}
Tak~Yeon Lee, Alison Smith, Kevin Seppi, Niklas Elmqvist, Jordan {Boyd-Graber},
  and Leah Findlater. 2017.
\newblock The human touch: {{How}} non-expert users perceive, interpret, and
  fix topic models.
\newblock \emph{International Journal of Human-Computer Studies}, 105:28--42.

\bibitem[{Li et~al.(2017)Li, Miller, Chopra, Ranzato, and
  Weston}]{liDialogueLearningHumanInTheLoop2017}
Jiwei Li, Alexander~H. Miller, Sumit Chopra, Marc'Aurelio Ranzato, and Jason
  Weston. 2017.
\newblock \href {http://arxiv.org/abs/1611.09823} {Dialogue {{Learning With
  Human}}-{{In}}-{{The}}-{{Loop}}}.
\newblock \emph{arXiv:1611.09823 [cs]}.

\bibitem[{Liu et~al.(2018)Liu, Tur, {Hakkani-Tur}, Shah, and
  Heck}]{liuDialogueLearningHuman2018a}
Bing Liu, Gokhan Tur, Dilek {Hakkani-Tur}, Pararth Shah, and Larry Heck. 2018.
\newblock \href {http://arxiv.org/abs/1804.06512} {Dialogue {{Learning}} with
  {{Human Teaching}} and {{Feedback}} in {{End}}-to-{{End Trainable
  Task}}-{{Oriented Dialogue Systems}}}.
\newblock \emph{arXiv:1804.06512 [cs]}.

\bibitem[{Lo and Lim(2020)}]{loInteractiveEntityLinking2020}
Pei-Chi Lo and Ee-Peng Lim. 2020.
\newblock Interactive {{Entity Linking Using Entity}}-{{Word Representations}}.
\newblock In \emph{Proceedings of the 43rd {{International ACM SIGIR
  Conference}} on {{Research}} and {{Development}} in {{Information
  Retrieval}}}, pages 1801--1804, {Virtual Event China}. {ACM}.

\bibitem[{Rao and Daum{\'e}~III(2018)}]{raoLearningAskGood2018}
Sudha Rao and Hal Daum{\'e}~III. 2018.
\newblock \href {http://arxiv.org/abs/1805.04655} {Learning to {{Ask Good
  Questions}}: {{Ranking Clarification Questions}} using {{Neural Expected
  Value}} of {{Perfect Information}}}.
\newblock \emph{arXiv:1805.04655 [cs]}.

\bibitem[{Schoch et~al.(2020)Schoch, Yang, and Ji}]{schochThisProblem2020}
Stephanie Schoch, Diyi Yang, and Yangfeng Ji. 2020.
\newblock {This is a Problem, Don’t You Agree?” Framing and Bias in Human
  Evaluation for Natural Language Generation}.

\bibitem[{Settles(2011)}]{settlesClosingLoopFast}
Burr Settles. 2011.
\newblock \href {https://www.aclweb.org/anthology/D11-1136} {Closing the loop:
  Fast, interactive semi-supervised annotation with queries on features and
  instances}.
\newblock In \emph{Proceedings of the 2011 Conference on Empirical Methods in
  Natural Language Processing}, pages 1467--1478, Edinburgh, Scotland, UK.
  Association for Computational Linguistics.

\bibitem[{Simard et~al.(2014)Simard, Chickering, Lakshmiratan, Charles, Bottou,
  Suarez, Grangier, Amershi, Verwey, and Suh}]{simardICEEnablingNonExperts}
Patrice~Y. Simard, David~Maxwell Chickering, Aparna Lakshmiratan, Denis~Xavier
  Charles, L{\'{e}}on Bottou, Carlos Garcia~Jurado Suarez, David Grangier,
  Saleema Amershi, Johan Verwey, and Jina Suh. 2014.
\newblock \href {http://arxiv.org/abs/1409.4814} {{ICE:} enabling non-experts
  to build models interactively for large-scale lopsided problems}.
\newblock \emph{CoRR}, abs/1409.4814.

\bibitem[{Smith et~al.(2018)Smith, Kumar, {Boyd-Graber}, Seppi, and
  Findlater}]{smithClosingLoopUserCentered2018}
Alison Smith, Varun Kumar, Jordan {Boyd-Graber}, Kevin Seppi, and Leah
  Findlater. 2018.
\newblock Closing the {{Loop}}: {{User}}-{{Centered Design}} and {{Evaluation}}
  of a {{Human}}-in-the-{{Loop Topic Modeling System}}.
\newblock In \emph{Proceedings of the 2018 {{Conference}} on {{Human
  Information Interaction}}\&{{Retrieval}} - {{IUI}} 18}, pages 293--304,
  {Tokyo, Japan}. {ACM Press}.

\bibitem[{Stiennon et~al.(2020)Stiennon, Ouyang, Wu, Ziegler, Lowe, Voss,
  Radford, Amodei, and Christiano}]{stiennonLearningSummarizeHuman2020}
Nisan Stiennon, Long Ouyang, Jeff Wu, Daniel~M. Ziegler, Ryan Lowe, Chelsea
  Voss, Alec Radford, Dario Amodei, and Paul Christiano. 2020.
\newblock \href {http://arxiv.org/abs/2009.01325} {Learning to summarize from
  human feedback}.
\newblock \emph{arXiv:2009.01325 [cs]}.

\bibitem[{Trivedi et~al.(2019)Trivedi, Dadashzadeh, Handzel, Chapman,
  Visweswaran, and Hochheiser}]{trivedi2019interactive}
Gaurav Trivedi, Esmaeel~R Dadashzadeh, Robert~M Handzel, Wendy~W Chapman, Shyam
  Visweswaran, and Harry Hochheiser. 2019.
\newblock Interactive nlp in clinical care: Identifying incidental findings in
  radiology reports.
\newblock \emph{Applied clinical informatics}, 10(4):655.

\bibitem[{Wallace et~al.(2019)Wallace, Rodriguez, Feng, Yamada, and
  {Boyd-Graber}}]{wallaceTrickMeIf2019}
Eric Wallace, Pedro Rodriguez, Shi Feng, Ikuya Yamada, and Jordan
  {Boyd-Graber}. 2019.
\newblock Trick {{Me If You Can}}: {{Human}}-in-the-{{Loop Generation}} of
  {{Adversarial Examples}} for {{Question Answering}}.
\newblock \emph{Transactions of the Association for Computational Linguistics},
  7:387--401.

\end{thebibliography}
\bibliographystyle{acl_natbib}

\end{document}